\documentclass[10pt,twocolumn,letterpaper]{article}
\usepackage{titling}
\usepackage{ijcb}
\usepackage{times}
\usepackage{epsfig}
\usepackage{amsmath}
\usepackage{amssymb}
\usepackage{booktabs}
\usepackage{diagbox}
\usepackage{multirow,tabularx}

\usepackage[pagebackref,breaklinks,colorlinks]{hyperref}
\usepackage{float}
\usepackage{wrapfig}
\usepackage{multicol}
\usepackage{flushend}
\usepackage{balance}
\usepackage{enumitem}

\ijcbfinalcopy 
\ijcbfinalcopy

\begin{document}

\title{Does lossy image compression affect racial bias within face recognition?}

\author{
    Seyma Yucer,
    Matt Poyser,
    Noura Al Moubayed,
    Toby P. Breckon       \\
    Department of Computer Science,
  Durham University \\ Durham, UK
    }

\maketitle
\thispagestyle{empty}
\begin{abstract}

\noindent
\textbf{Yes} - This study investigates the impact of commonplace lossy image compression on face recognition algorithms with regard to the racial characteristics of the subject. We adopt a recently proposed racial phenotype-based bias analysis methodology to measure the effect of varying levels of lossy compression across racial phenotype categories. Additionally, we determine the relationship between chroma-subsampling and race-related phenotypes for recognition performance. Prior work investigates the impact of lossy JPEG compression algorithm on contemporary face recognition performance. However, there is a gap in how this impact varies with different race-related inter-sectional groups and the cause of this impact. Via an extensive experimental setup, we demonstrate that common lossy image compression approaches have a more pronounced negative impact on facial recognition performance for specific racial phenotype categories such as darker skin tones (by up to 34.55\%). Furthermore, removing chroma-subsampling during compression improves the false matching rate (up to 15.95\%) across all phenotype categories affected by the compression, including darker skin tones, wide noses, big lips, and monolid eye categories. In addition, we outline the characteristics that may be attributable as the underlying cause of such phenomenon for lossy compression algorithms such as JPEG.
\end{abstract}
\vspace{-0.3cm}
\section{Introduction}
\label{sec:intro}

A growing number of studies focus on racial bias within face recognition due to the prevalence of disparate real-world performance on inter-sectional racial groups \cite{nist_demographic}. Such attention has forced several organisations to withdraw their algorithms or datasets due to racial biases, and disparities \cite{shiaeles2021facebook,menn2019microsoft,castelvecchi2020facial}. Nevertheless, there are still many areas, such as employment, public security, criminal justice, and credit reporting, where face recognition applications are in use, meaning that we need fair, trustworthy, and bias-free face recognition \cite{nist_verification, nist_identification}.

From image acquisition to evaluation, all phases of face recognition are prone to bias. However, most research focuses on the latter aspects of dataset collection and model evaluation to explore and mitigate such bias \cite{serna2022sensitive,wang2021meta,gong2021mitigating}. As such, many datasets and annotations have been released \cite{sixta2020fairface,wang2020mitigating}, generative adversarial networks have been explored to enrich under-represented groups during training \cite{georgopoulos2021mitigating,yucer2020exploring} and regularisation methods have been proposed to minimise performance differences between subgroups \cite{tartaglione2021end}. Furthermore specific evaluation methodologies have been devised to tackle bias collaboratively \cite{joshi2022fair,cavazos2020accuracy,yucer2022measuring}. Despite this plethora of research, no studies examine the potential impact of image acquisition decisions when addressing racial bias within face recognition. Any source of bias at this early stage is just propagated and exacerbated within contemporary face recognition approaches \cite{mehrabi2021survey}.

On the other hand, existing image acquisition standards for face recognition systems such as ISO/IEC 19794-5 \cite{iso19} and ICAO 9303 \cite{monnerat2007machine} propose both image-based (i.e. illumination, occlusion) and subject-based (i.e. pose, expression, accessories) quality standards to ensure facial image quality. Accordingly, facial images should also be stored using lossy image compression standards such as JPEG \cite{JPEG} or JPEG2000 \cite{JPEG2000}; and identifiable for gender, eye colour, hair colour, expression, properties (i.e. glasses), pose angles (yaw, pitch, and roll), and landmark positions. However, common face recognition benchmarks do not conform to the ISO/IEC 19794-5 and ICAO 9303 standards. Moreover, in-the-wild samples are often obtained under the varying camera and environmental conditions to challenge the proposed solutions. Nevertheless, most facial image samples within such datasets are compressed via lossy JPEG compression \cite{pennebaker1992jpeg}.

Accordingly, some limited previous work \cite{dodge2016understanding,vasiljevic2016examining,koziarski2018impact} focuses on the impact of low-quality, blurred, noisy or distorted imagery on Convolutional Neural Network (CNN) based image recognition or classification. Dodge and Karam \cite{dodge2017study} highlight a significant decrease in contemporary neural network performance, whilst human examiners remain resilient to such factors. Particularly, Torfason \cite{torfason2018towards} focuses on compression methods and bypasses the decoding phase of image compression. They point out that encoded representations are more advantageous than compressed/decoded images for classification and semantic segmentation. Poyser \cite{poyser2021impact} evaluates the impact of lossy compression algorithms on various CNN architectures, in which they measure the robustness and performance impact of compression for various computer vision tasks. They determine that, in general, CNN architectures can be resilient to the introduction of lossy JPEG compression artefacts if the initial training regime includes the use of compressed images \cite{poyser2021impact}. These results align with the findings of Zanjani \cite{zanjani}, who considers the impact of JPEG 2000 compression \cite{JPEG2000} on CNN for cancer diagnosis systems. Indeed, retraining the CNN architecture on lossily compressed images affords a 59\% performance increase for tumour detection within compressed test imagery \cite{zanjani}.

For face recognition approaches, the National Institute of Standards and Technology (NIST) provides a comprehensive assessment of compressed image influence on facial recognition algorithms \cite{quinn2011performance}. It investigates the speed versus accuracy trade-off for early machine learning-based face recognition algorithms. Karahan \cite{karahan2016image} indicates that image blur, noise, and occlusion can cause significant degradation in face recognition accuracy. Another study \cite{yang2018quality} improves face detection and recognition performance on low-quality images by introducing a fusion quality prediction network. Moreover, Terh\"orst \cite{terhorst2020face} shows quality assessment algorithms are skewed towards the subgroups which are also affected by face recognition bias. 

Prior literature on image acquisition operations (compression, quality assessment) for face recognition \cite{hernandez2020biometric} are limited with regard to racial bias and its race-based phenotypic influence, which is where this study is focused. The most related work to ours, \cite{majumdar2021unravelling} explores the test image distortion impact on pre-trained face recognition models using binary gender G1 (Male) and G2 (Female), and race R1 (light skin colour) and R2 (dark skin colour) subgroups. As a result, they find that the regions of interest used in the models shift towards less discriminatory regions in the presence of distortions, resulting in unequal performance degradation among subgroups.

In this study, we examine whether lossy image compression adversely impacts phenotype-based racial performance bias within face recognition during training and testing. We estimate such impact on phenotype attribute categories individually. Furthermore, we also investigate differing chroma-subsampling rates to assess how this common lossy compression colour-related trait directly impacts recognition performance across varying phenotype-based categories. More precisely, however, we determine the relationship between the level of compression and chroma-subsampling applied and recognition performance in order to allow us to build a better understanding. 

\noindent
To these ends, we adapt the recently established evaluation methodology \cite{yucer2022measuring} that introduces phenotype-based racial bias measurement for face recognition. Furthermore, we determine the effect of varying factors, including the compression levels of lossy JPEG \cite{JPEG} image encoding, chroma-subsampling, and compressed versus non-compressed training on different race-based phenotype categories in order to evaluate the racial bias across multiple face recognition datasets. In this paper, our key contributions are as follows:

\noindent
\begin{itemize}[leftmargin=*]
\item we evaluate the impact of lossy image compression on CNN-based facial recognition approaches across different racial characteristics using the phenotype-based methodology \cite{yucer2022measuring}, extending the earlier studies of \cite{yucer2022measuring, hernandez2020biometric, poyser2021impact}.

\vspace{-0.2cm}
\item we compare several variants of training strategies, including lossy compression, within the balanced/imbalanced training datasets and race-related facial phenotypes.

\vspace{-0.2cm}
\item we experimentally demonstrate that the use of lossy image compression during inference adversely affects the performance of contemporary face recognition approaches \cite{deng2019arcface} on a subset of race-related facial phenotype grouping (i.e. darker skin tones, monolid eye shape) and that its effect is present regardless of whether compressed imagery is used for model training.

\vspace{-0.2cm}
\item we investigate the specific impact of chroma-subsampling on bias performance by comparing recognition performance with and without chroma-subsampling within lossy compressed facial imagery.

\end{itemize}
\section{Experimental Methodology}
\noindent

\begin{table}[t]

\setlength\tabcolsep{10pt}
\resizebox{\columnwidth}{!}{
\begin{tabular}{l|c}
\toprule
\textbf{Attribute} & \textbf{Categories} \\
\midrule
\textbf{Skin Type} & Type 1 / 2 / 3 / 4 / 5 / 6 \\
\textbf{Eyelid Type} & Monolid / Other\\
\textbf{Nose Shape} & Wide / Narrow \\
\textbf{Lip Shape} & Full / Small \\
\textbf{Hair Type} & Straight / Wavy / Curly / Bald \\
\textbf{Hair Colour} & Red / Blonde / Brown / Black / Grey \\ 
\bottomrule
\end{tabular}
}
\vspace{0.1cm}
\caption{Adapted Facial phenotype attributes and their categorisation from \cite{yucer2022measuring}.}
\label{tab:cat}
\vspace{-0.5cm}
\end{table}

\begin{figure}[t]
 \centering
 \includegraphics[width=0.9\linewidth]{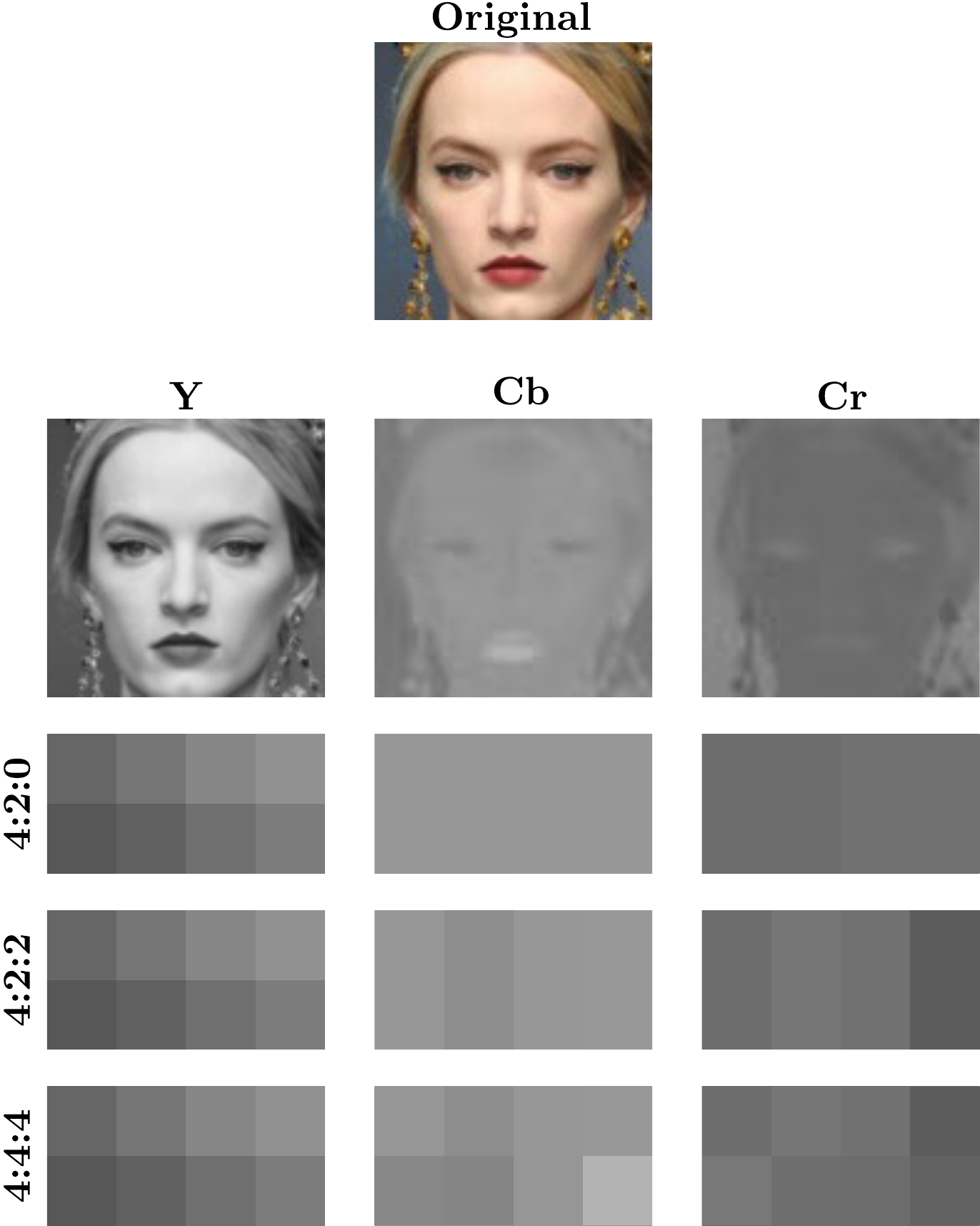}
 \caption{Chroma subsampling operation on different rates (4:2:0, 4:2:2, 4:4:4). Each rate differs according to how many pixels will be the same in the block.}\label{fig:cd}
\vspace{-0.6cm}
\end{figure}

\noindent
In this section, we explain the phenotype-based racial bias evaluation methodology used (Section \ref{sec:wacv}), the most widespread lossy image compression process (JPEG, Section \ref{sec:ic}), how we evaluate the influence of chroma subsampling (Section \ref{sec:cd}), our compression level selection (Section \ref{sec:complev}), and the training strategies used (Section \ref{sec:ts}) for the generation of our results (Section \ref{sec:res}). 

\subsection{Phenotype-based Bias Analysis Methodology}
\label{sec:wacv}

\noindent
Previous work highlights the negative impacts of using standard geographically based racial grouping labels to evaluate cross-race face recognition performance \cite{hanna2020towards,raji2020saving}. Accordingly, many studies \cite{yucer2022measuring,raji2020saving,scheuerman} suggest avoiding using erroneous racial or binary skin tones grouping strategies or exposing protected demographic attributes that can cause privacy and consent violations for individuals.

Alternatively, we adopt a racial bias analysis methodology that uses facial phenotype attributes for face verification (one-to-one facial matching) task \cite{yucer2022measuring}. The study categorises representative racial characteristics on the face and audits these attributes: skin types, eyelid type, nose shape, lips shape, hair colour and hair type for two different publicly available face datasets: VGGFace2 (test set) \cite{cao2018vggface2}, and RFW \cite{wang2019racial}. We show each of the predefined phenotypes and their categories in Table \ref{tab:cat}. Moreover, this methodology provides different pairing strategies for face verification to draw attention to the importance of pairing for comprehensive evaluation. It introduces attribute-based pairings, which contain same-attribute grouping pair combinations to compare individual attribute performance for face verification. Additionally, the study shares cross-attribute pairing combinations between each grouping to measure false matching rates between all possible attribute category pair combinations.

         
         

\begin{figure*}[ht]
     \centering
         \includegraphics[width=\textwidth]{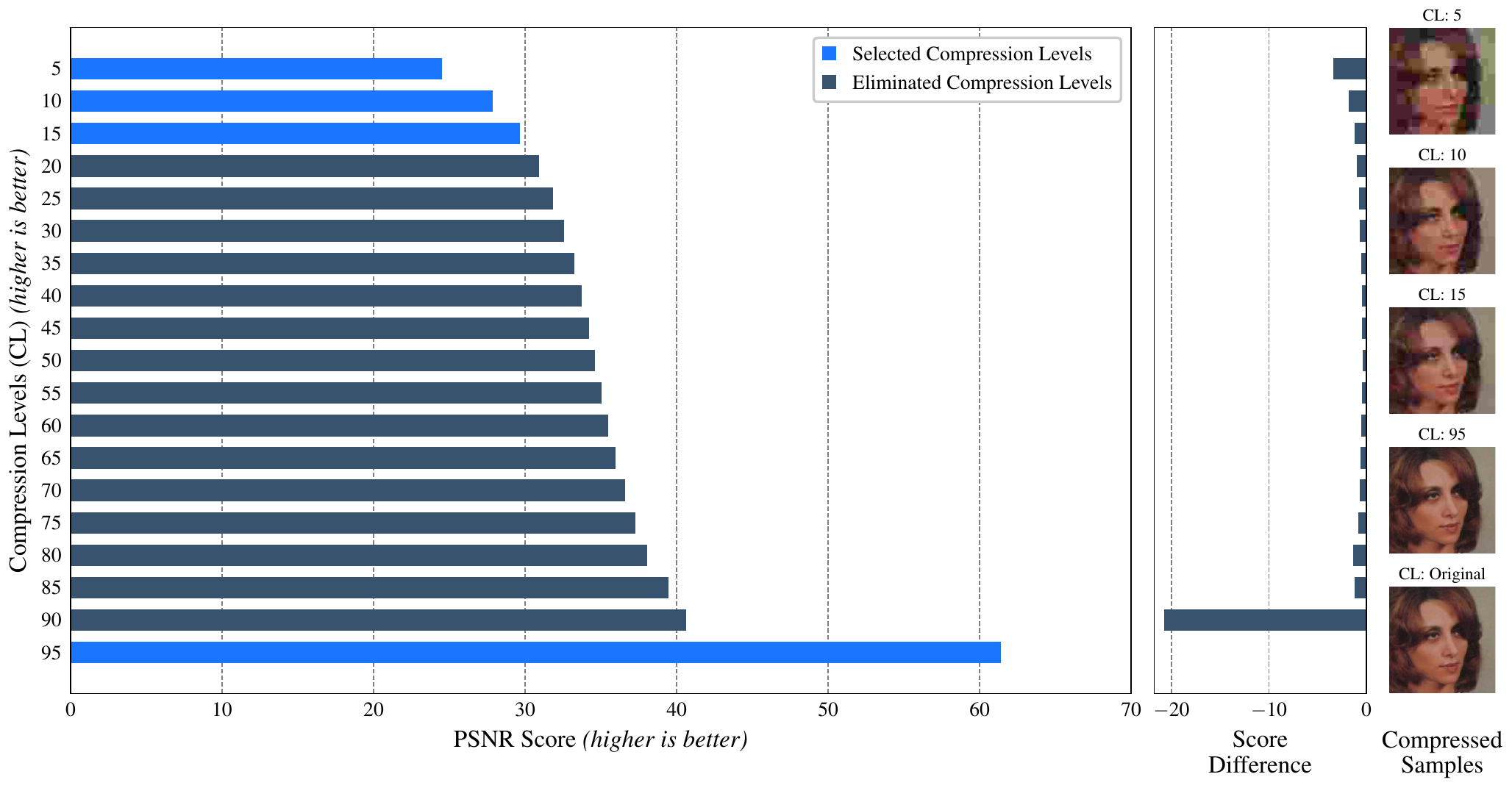}
         \caption{PSNR Scores of RFW dataset on different compression levels (CL). Relative score difference shows how much the quality changes at each level. \label{compression_levels}}
\vspace{-0.5cm}
\end{figure*}

On this basis, we use the set of observable characteristics of an individual face where race-related facial phenotype labels provide a relation between the task performance for a given face image sample under varying levels of lossy image compression and its underlying racial characteristics.

\subsection{Lossy Image Compression}
\label{sec:ic}
\noindent
The Joint Photographic Experts Group (JPEG), an international image compression standard \cite{JPEG} for still images, operates within manageable algorithmic space and time complexity whilst offering good reconstruction image quality. The JPEG standard defines four operating modes \textit{(1: Sequential Lossless Mode, 2: Sequential DCT-based Mode, 3: Progressive DCT-based Mode, 4: Hierarchical Mode)}, formed by an encoder and decoder which follow block-based transform coding. The image encoding strategy includes colour space transformation (from RGB to YCrCb), chroma channel subsampling, Discrete Cosine Transform (DCT), quantisation and entropy coding to compress the image \cite{JPEG}.

In this study, we use ImageMagick Library (version 7.0.11.13) to perform JPEG compression (via libjpeg 8). The implementation switches the JPEG operational modes according to the compression level specified (i.e. quality level $q$, range: 0 - 100 for JPEG, higher = better image quality, less information loss $+$ larger file sizes). Similar to the mode one operation, it does not down-sample the chroma channels if the compression level is higher than 90 (i.e. there is no colour-based information loss for compression, $q=90$). It applies the baseline JPEG algorithm between compression levels 90 and 10, which is sequential DCT-based Mode (2). For compression levels, $(q=90)$, lossy compression is applied to both the luminance channel, Y, and the colour containing chroma channels, ${Cr, Cb}$.

\subsection{Chroma Subsampling}
\label{sec:cd}

\noindent
Standard lossy compression algorithms such as JPEG contain a colour space reduction step, as the human eye is less sensitive to chromatic (i.e.  colour) changes than changes in illumination (i.e. brightness). In this step, the luminance channel (Y) remains unchanged, but the image colour space (Cr and Cb) is reduced. Subsequently, by default JPEG algorithm employs 4:2:0 chroma subsampling to reduce the colour information of the original image. It takes a 2-by-2-pixel block within each block and assigns the same colour (the colour of the top-left pixel) while the luminance component varies. Alternatively, for less colour information reduction, 4:2:2 with half sampling rate horizontally takes 2 pixels in each row and assigns the same colour. In Figure \ref{fig:cd}, we illustrate the three different sampling ratios (4:2:0, 4:2:2 and 4:4:4 no subsampling) on image pixels. In this first step of compression, chroma subsampling converts the image to YCbCr colour space and then reduces the chroma channels ${Cb, Cr}$ information by assigning the top-left block pixel value to other pixels in the block. Block size and how many pixel values remain vary according to the sampling ratio. 

This evaluation investigates the effect of sampling ratio on phenotype-based face recognition performance. We compare the default 4:2:0 subsampling with the 4:4:4 no chroma-subsampling factor, which keeps luminance and colour information in its entirety (i.e. unchanged). The rationale behind this evaluation is that if chroma subsampling has a profound impact on recognition performance, we can avoid this issue by recommending the use of 4:4:4 (no chroma-subsampling) with only a small impact on compression performance.

\subsection{Compression Level Selection}
\label{sec:complev}

\noindent
In order to ascertain the impact of lossy compression on face recognition performance, we are interested in the resulting reduction in image quality at varying levels of JPEG compression.
Consequently, we analyse uniformly distributed compression levels on the RFW benchmark face recognition dataset \cite{wang2019racial} using PSNR; Peak signal-to-noise ratio \cite{mittal2012no}. PSNR score is correlated with the quality of reconstruction of lossy JPEG compression. In Figure \ref{compression_levels}, we show the relation between the PSNR  score versus the JPEG compression level, $q$. Firstly, we uniformly select levels $q=\{5...95\}$ in intervals of 5 and compress the whole dataset to each of these JPEG compression levels. Secondly, we measure the PSNR score on all levels and highlight the relative score difference. Based upon this analysis, we downselect the set of JPEG compression levels $(q= \{5, 10, 15\})$, in which quality decrease is most apparent (PSNR score decreases harshly). In addition, we select $q=95$ as it represents the case where there is no chroma down-sampling used within the lossy compression scheme.

\subsection{Training Strategies}
\label{sec:ts}
\noindent
We design different test scenarios to measure the impact of image compression on face verification performance. 

\noindent
\textbf{Racially Imbalanced Dataset:} Firstly, we train ArcFace \cite{deng2019arcface} with ResNet101v2 \cite{he2016deep} on the original aligned VGGFace2 benchmark dataset \cite{cao2018vggface2}, containing 3.3 million images with 8631 subjects where subject distribution is racially imbalanced. Subsequently, we test using the RFW benchmark dataset \cite{wang2019racial} with the original (aligned) images and compressed images to each of the previously down-selected JPEG compression levels. We then repeat the training on the VGGFace2 benchmark dataset \cite{cao2018vggface2} four times, having first compressed the entire dataset to each of the down-selected JPEG compression levels. This results in four ArcFace models, each trained on image samples at a different JPEG compression level. Subsequently, we measure the performance of each of these four trained ArcFace models using the RFW benchmark dataset \cite{wang2019racial} that has been compressed to the corresponding JPEG compression level upon which each of the models was trained.
\begin{figure*}
 \centering
 \includegraphics[width=1\linewidth]{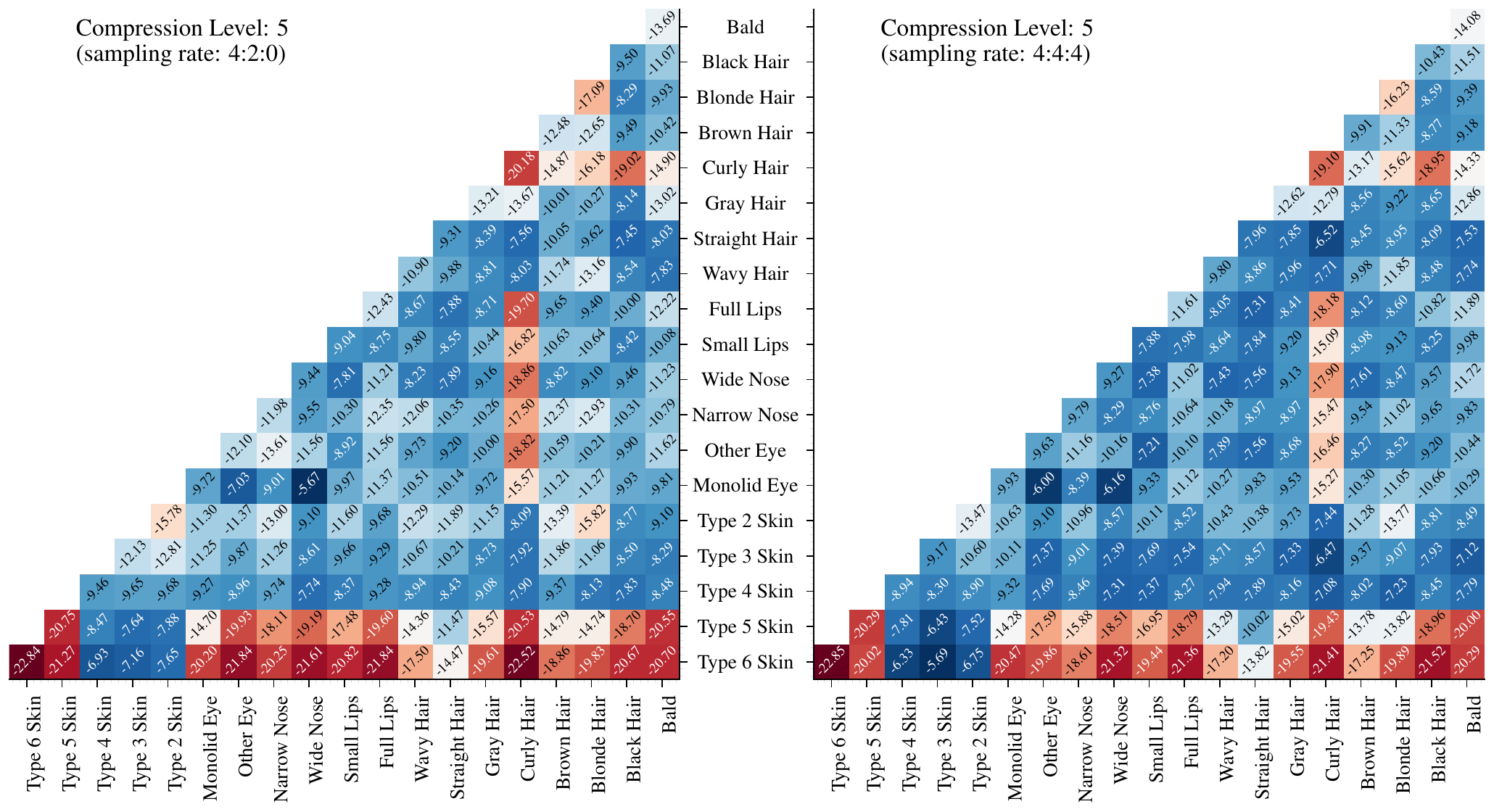}
 \caption{BUPT-Balanced non-compressed, compressed RFW test imagery (q=5); FMR performance differences of cross-attribute based pairings. Each cell depicts $FMR_{original}-FMR_{q}$.\label{fig:set1}}
 \vspace{-0.1cm}
\end{figure*}

\noindent
\textbf{Racially Balanced Dataset:} 
Similar to the imbalanced train set strategy, we train ArcFace \cite{deng2019arcface} with ResNet50 on the original aligned BUPT-Balanced benchmark dataset \cite{wang2020mitigating} that contains 28000 face subjects containing balanced racial distributions among four groups \textit{\{African, Asian, Indian, Caucasian\}} with 7000 subjects each. Subsequently, we repeat the training on the BUPT-Balanced benchmark dataset \cite{wang2020mitigating} four times, having first compressed the entire dataset to each of the same down-selected JPEG compression levels. This way, another four ArcFace models are trained on image samples at a different JPEG compression level. Additionally, we replicate non-compressed and compressed training at level 5 $(q=5)$ by removing chroma subsampling (4:4:4) to measure the impact of the colour reduction step in lossy compression on face verification performance.

\section{Results and Discussion}
\noindent
This section provides extensive experimental results to understand the impact of chroma subsampling and compressed training imagery using two different dataset training datasets and different compression levels. Additionally, we place extended results in Supplementary Material.
\label{sec:res}

\begin{figure*}
 \centering
 \includegraphics[width=1\linewidth]{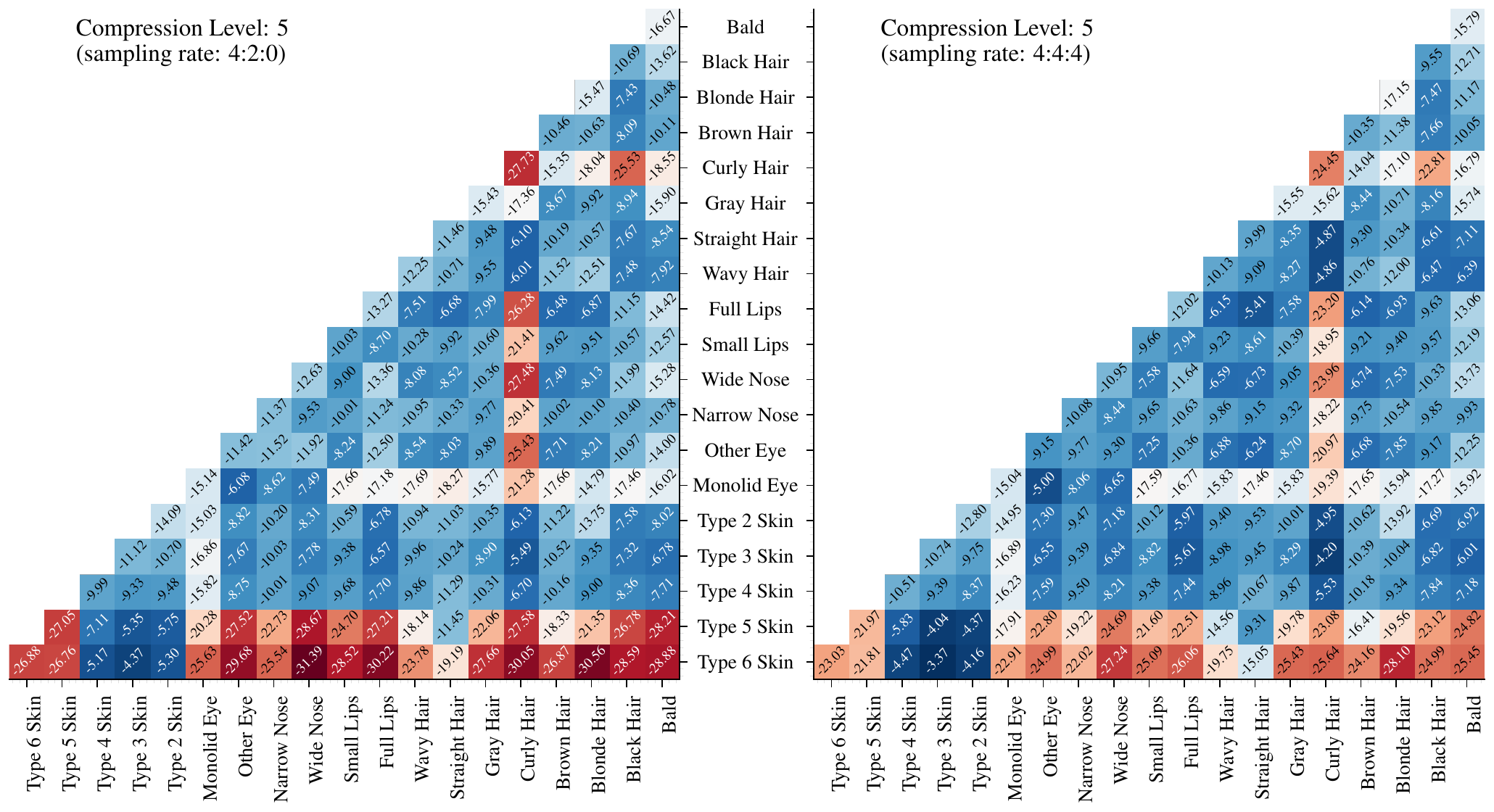}
 \caption{VGGFace2 non-compressed, compressed RFW test imagery $(q=5)$; FMR performance differences of cross-attribute based pairings. Each cell depicts $FMR_{original}-FMR_{q}$.\label{fig:set2}}

\end{figure*}

\begin{figure*}
 \centering
 \includegraphics[width=1\linewidth]{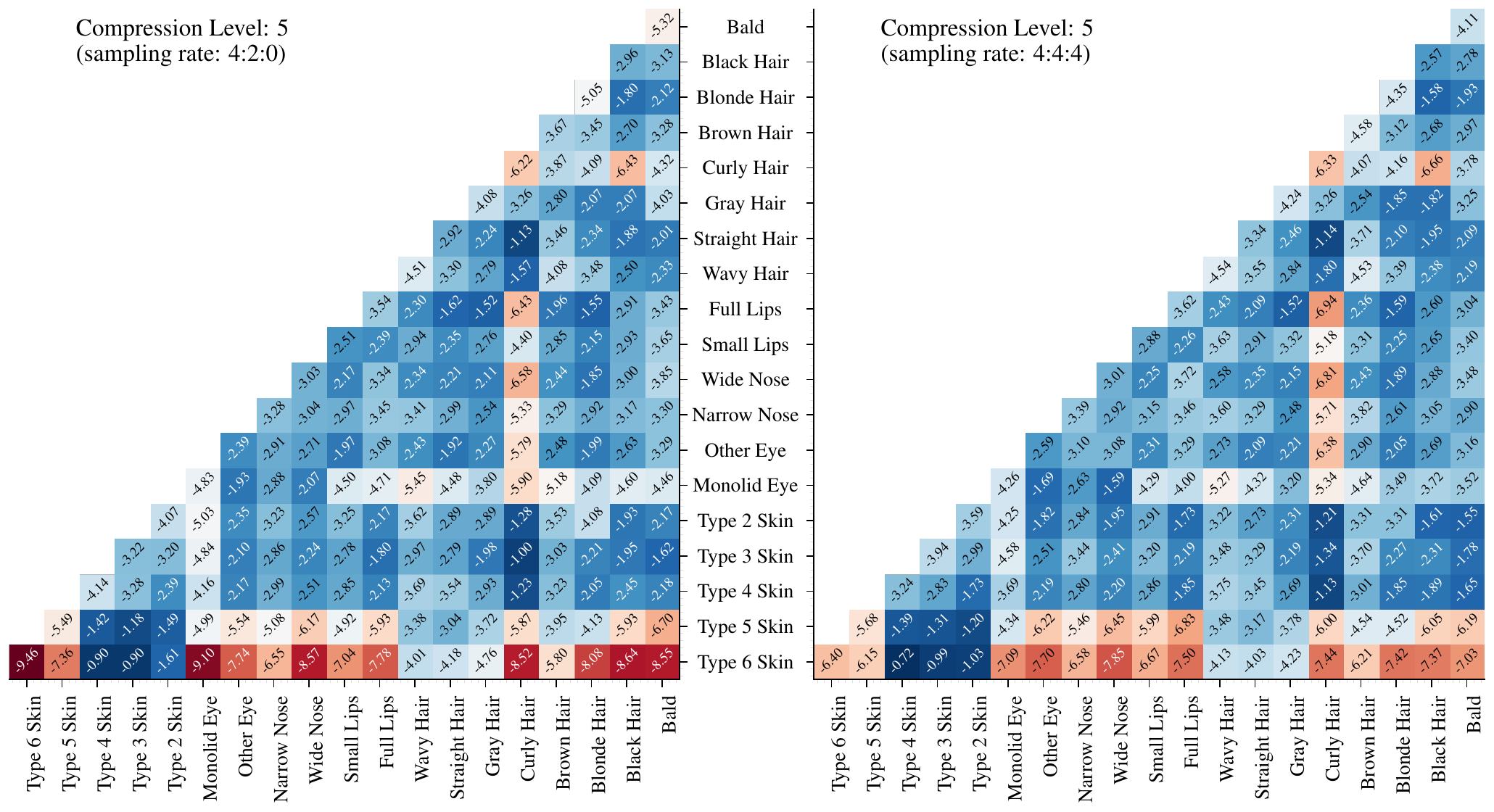}
 \caption{BUPT-Balanced compressed $(q=5)$, compressed RFW test imagery $(q=5)$; FMR performance differences of cross-attribute based pairings. Each cell depicts $FMR_{original}-FMR_{q}$.\label{fig:set3}}

\end{figure*}

\subsection{False Verification Matching Rates}

\noindent
In this section, we present False Matching Rate (FMR) differences for each of the proposed training strategies in Section \ref{sec:ts} and the down-selected compression levels (Figure \ref{compression_levels}). FMR is a critical metric, such that any change in performance may result in false facial verification and the associated consequences  \cite{nist_verification}.

Figures \ref{fig:set1}, \ref{fig:set2}, \ref{fig:set3} show the FMR changes under the varying sampling rates of lossy image compression and how this varies across the racial phenotype labels associated with the dataset. Using the cross attribute pairings provided by \cite{yucer2022measuring}, we evaluate $FMR_{original}-FMR_q$ where $FMR_{original}$ is FMR of non-compressed training and test imagery. $FMR_q$ is the FMR of compressed or non-compressed training but compressed test imagery at down-selected level $q$. Smaller (and negative) values indicate a more considerable decline from the original level of performance.

\noindent
\textbf{Compression Levels:} We observe that for all down-selected compression levels $q= \{5, 10, 15, 95\}$, the FMR increases when additional lossy compression is applied, demonstrating that compression level 5 (the highest compression rate) results in the most significant decrease in FMR performance, whilst compression level 95 (the lowest compression rate) does not result in any noticeable FMR performance differences. We compare compression levels 95, 15, 10 and 5 with baseline results to show how FMR rise at higher compression levels. For additional performance results on different levels, see Supplementary Materials.



\begin{table*}[t!]
\setlength{\tabcolsep}{11pt}
\centering
\begin{tabular}{@{}lccccccccc@{}}
\toprule
\multicolumn{1}{c}{}               & \multicolumn{4}{c}{None-Compressed Training Set}                                                                   & \multicolumn{4}{c}{Compressed Training Set}                                                                        &          \\ \midrule
\multicolumn{1}{l}{Attribute Name} & 95                        & 15                        & 10                        & 5                          & 95                        & 15                        & 10                        & 5                          & Original \\ \midrule
\multicolumn{1}{l|}{Curly Hair}    & 93.10                     & 82.37                     & 75.80                     & \multicolumn{1}{c|}{59.53} & 92.77                     & 87.20                     & 82.90                     & \multicolumn{1}{c|}{73.27} & 93.15    \\
\multicolumn{1}{l|}{Full Lips}      & 93.37                     & 83.55                     & 77.03                     & \multicolumn{1}{c|}{61.37} & 92.80                     & 87.97                     & 83.62                     & \multicolumn{1}{c|}{75.30} & 93.38    \\
\multicolumn{1}{l|}{Monolid Eye}   & 93.25                     & 83.43                     & 77.28                     & \multicolumn{1}{c|}{63.18} & 93.48                     & 87.62                     & 85.10                     & \multicolumn{1}{c|}{76.95} & 93.30    \\
\multicolumn{1}{l|}{Type 5}        & 94.87                     & 85.98                     & 80.17                     & \multicolumn{1}{c|}{60.32} & 94.53                     & 90.22                     & 87.03                     & \multicolumn{1}{c|}{76.97} & 94.85    \\
\multicolumn{1}{l|}{Type 6}        & 94.85                     & 86.55                     & 79.35                     & \multicolumn{1}{c|}{61.75} & 94.43                     & 90.02                     & 86.20                     & \multicolumn{1}{c|}{77.72} & 94.82    \\
\multicolumn{1}{l|}{Black Hair}    & 93.70                     & 85.13                     & 79.97                     & \multicolumn{1}{c|}{65.83} & 93.50                     & 89.55                     & 86.87                     & \multicolumn{1}{c|}{77.92} & 93.73    \\
\multicolumn{1}{l|}{Wide Nose}     & 93.95                     & 85.53                     & 79.97                     & \multicolumn{1}{c|}{63.15} & 93.42                     & 89.57                     & 86.78                     & \multicolumn{1}{c|}{78.33} & 93.98    \\
\multicolumn{1}{l|}{Other Eye}     & 94.32                     & 86.65                     & 81.10                     & \multicolumn{1}{c|}{65.28} & 93.70                     & 89.57                     & 87.43                     & \multicolumn{1}{c|}{78.55} & 94.38    \\
\multicolumn{1}{l|}{Type 4}        & \multicolumn{1}{l}{94.05} & \multicolumn{1}{l}{87.72} & \multicolumn{1}{l}{83.47} & \multicolumn{1}{l|}{67.28} & \multicolumn{1}{l}{93.72} & \multicolumn{1}{l}{89.67} & \multicolumn{1}{l}{87.45} & \multicolumn{1}{l|}{79.23} & 94.07    \\
\multicolumn{1}{l|}{Type 1}        & 92.86                     & 86.88                     & 84.72                     & \multicolumn{1}{c|}{72.43} & 94.19                     & 89.87                     & 88.21                     & \multicolumn{1}{c|}{79.57} & 92.86    \\
\multicolumn{1}{l|}{Straight Hair} & 94.18                     & 86.70                     & 81.98                     & \multicolumn{1}{c|}{66.15} & 93.92                     & 89.43                     & 86.28                     & \multicolumn{1}{c|}{79.65} & 94.12    \\
\multicolumn{1}{l|}{Narrow Nose}   & 94.35                     & 86.30                     & 80.07                     & \multicolumn{1}{c|}{66.73} & 94.60                     & 89.63                     & 87.20                     & \multicolumn{1}{c|}{79.77} & 94.43    \\
\multicolumn{1}{l|}{Type 3}        & 94.05                     & 86.07                     & 81.03                     & \multicolumn{1}{c|}{67.05} & 94.32                     & 89.48                     & 86.80                     & \multicolumn{1}{c|}{79.93} & 93.98    \\
\multicolumn{1}{l|}{Small Lips}    & 94.35                     & 87.28                     & 82.03                     & \multicolumn{1}{c|}{67.53} & 95.00                     & 90.63                     & 87.97                     & \multicolumn{1}{c|}{81.22} & 94.37    \\
\multicolumn{1}{l|}{Wavy Hair}     & 95.87                     & 89.05                     & 84.63                     & \multicolumn{1}{c|}{69.53} & 95.52                     & 92.17                     & 89.33                     & \multicolumn{1}{c|}{82.73} & 95.83    \\
\multicolumn{1}{l|}{Brown Hair}    & 95.12                     & 88.40                     & 83.33                     & \multicolumn{1}{c|}{67.32} & 95.23                     & 91.85                     & 89.03                     & \multicolumn{1}{c|}{82.80} & 95.15    \\
\multicolumn{1}{l|}{Bald Hair}     & 96.55                     & 90.43                     & 85.93                     & \multicolumn{1}{c|}{67.62} & 95.88                     & 93.07                     & 90.37                     & \multicolumn{1}{c|}{83.13} & 96.55    \\
\multicolumn{1}{l|}{Red Hair}      & 96.91                     & 90.57                     & 84.97                     & \multicolumn{1}{c|}{71.20} & 96.33                     & 92.49                     & 89.98                     & \multicolumn{1}{c|}{84.89} & 96.91    \\
\multicolumn{1}{l|}{Type 2}        & 96.27                     & 89.98                     & 85.98                     & \multicolumn{1}{c|}{68.45} & 96.57                     & 94.27                     & 91.58                     & \multicolumn{1}{c|}{85.93} & 96.33    \\
\multicolumn{1}{l|}{Gray Hair}     & 96.53                     & 92.47                     & 88.83                     & \multicolumn{1}{c|}{72.60} & 96.42                     & 94.35                     & 91.93                     & \multicolumn{1}{c|}{86.75} & 96.55    \\
\multicolumn{1}{l|}{Blonde Hair}   & 97.15                     & 92.50                     & 88.52                     & \multicolumn{1}{c|}{71.55} & 97.15                     & 94.83                     & 93.40                     & \multicolumn{1}{c|}{87.85} & 97.15    \\ \midrule
\multicolumn{1}{l|}{Mean Accuracy} & 94.74                     & 87.31                     & 82.20                     & \multicolumn{1}{c|}{66.47} & 94.64                     & 90.64                     & 87.88                     & \multicolumn{1}{c|}{80.40} & 94.76    \\ \midrule
\multicolumn{1}{l|}{STD}           & 1.31                      & 2.76                      & 3.58                      & \multicolumn{1}{c|}{3.85}  & 1.27                      & 2.18                      & 2.61                      & \multicolumn{1}{c|}{3.81}  & 1.31     \\ \bottomrule
\end{tabular}

 \vspace{0.2cm}
\caption{Verification performance on RFW test set using uncompressed (left) and compressed (right) training imagery. Attribute-based pairings are those from the study of \cite{yucer2022measuring}.}
\vspace{-0.1cm}
\label{tab:acc_lev}

\end{table*}

\noindent
\textbf{Chroma subsampling vs No-chroma subsampling} We compress all the imagery in the BUPT-Balanced training dataset under two different sampling rates, 4:2:0 (JPEG default) and 4:4:4 on compression level 5 $(q=5)$. The FMR cross-attribute category results are compared in Figures \ref{fig:set1}, \ref{fig:set2}, \ref{fig:set3}. For non-compressed and compressed training, the 4:4:4 sampling rate decreases the FMR for all phenotype categories meaning that removing chroma sampling within the image encoding strategy of the lossy compression technique improves the performance difference and reduces the prevalence of the bias. Accordingly, we evaluate the average FMR for each phenotype category and calculate the standard deviation across all categories. Indeed, for both training strategies in Figure \ref{fig:set2} and \ref{fig:set3}, using no chroma-sampling improves FMR variation across all categories. For VGGFace2 non-compressed training (Figure \ref{fig:set2}), standard deviation drops from 3.91 to 3.28 $(15.95 \% \downarrow)$, whilst BUPT compressed training (Figure \ref{fig:set3}) standard deviation drops from from 0.91 to 0.81 $(10.88 \% \downarrow)$.

\noindent
\textbf{Non-compressed vs compressed training sets:} When the model is trained on original/non-compressed training imagery (Figures \ref{fig:set1} and \ref{fig:set2}), FMR on darker skin tone (Type 5-6) increases considerably compared to other phenotypes such as lighter skin tones (Types 2-4) with the introduction of lossy compression at test time. At the highest level of compression $(q=5)$, the increase in FMR is greater when both phenotype categories in the pair are correlated with the stereotypically African/Afro-Caribbean racial features \cite{zhuang2010facial}. For instance, the Full Lips $\leftrightarrow$ Type 6 pair has the highest FMR among all other pairs higher than Type 2 $\leftrightarrow$ Type 6 skin tone pairings. For compressed training imagery (Figures \ref{fig:set3} and Supplementary \ref{fig:sup_set3}), we observe improved results for both imbalanced and balanced dataset training. However, darker skin tone and related categories still maintain FMR higher than the other phenotype categories.


\noindent
\textbf{Racially balanced vs imbalanced training sets:} Using the racially balanced dataset for training does not ameliorate FMR differences among such pairings. For example, at the highest level of compression $(q=5)$, the average performance decrease of all skin tone Type 5 pairings (Type 5-Bald, Type 5-Black Hair etc.) is 16.06\% for imbalanced dataset training (Figure \ref{fig:set1}). At the same time, it is decreases by 17.69\% (Figure \ref{fig:set2}) from balanced dataset training. 
However, in racially imbalanced training, the FMR results for pairings with monolid eyes degrade more compared to racially balanced training. As there are significantly fewer monolid eye face samples than other phenotypes in the imbalanced VGGFace2 dataset, we assume that their representation degrades more than other phenotypes as the lossy compression level increases.

\subsection{Attribute-based Verification vs. Compression Levels}

\noindent
We additionally present attribute-based verification accuracy for the down-selected compression levels applied at training and test time for the BUPT-Balanced benchmark dataset \cite{wang2020mitigating}. Moreover, we provide supporting evidence of compressed \textit{vs.} uncompressed training set face verification performance in Table \ref{tab:acc_lev}. We use the same 6000 (3000 positive 3000 negatives) attribute-based image pairings provided by \cite{yucer2022measuring}. For both non-compressed and compressed training setups, we show that as the compression increases, the standard deviation across all phenotype categories increases (as a measure of non-uniform performance and bias). Similarly, accuracy decreases for all phenotype categories. However, using uncompressed training imagery (Table 2, left) results in a further decline in performance for darker skin tones Type 5-6, curly hair, full lips and monolid eye, when compared to other facial phenotypes, as the level of lossy compression within the test set is increased. Skin Type 5 attribute pairings accuracy drops from 94.87\% to 60.32\% $(34.55 \% \downarrow)$, while Skin Type 2 attribute accuracy drops from 96.33\% to 68.45\% $(27.88 \% \downarrow)$.  Similar to the non-compressed training set, we do observe non-uniform disparate changes in accuracy when the model is trained on compressed imagery (Table \ref{tab:acc_lev}, right). Furthermore, the compressed training set produces a smaller standard deviation in accuracy between phenotype categories.

\begin{figure}[t]
 \includegraphics[width=1\linewidth]{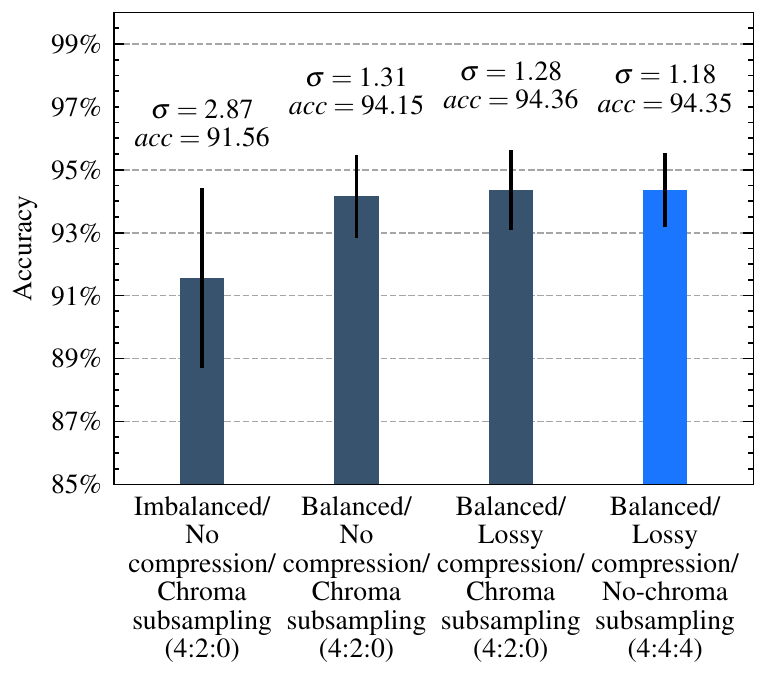} 
 \vspace{-0.25cm}
 \caption{Mean Accuracy and standard deviation of all attribute categories and their comparison on different training strategies using compressed $(q=75)$ RFW test set.  \label{fig:final_comp}
\vspace{-0.60cm}}
 
\end{figure}

Lastly, we summarise the relationship between all factors (dataset distribution, compression, chroma subsampling) in Figure \ref{fig:final_comp}. We evaluate attribute-based pairings accuracy for all phenotype categories and compare different training strategies mean accuracy and standard deviations. We change one factor during training in each strategy and provide corresponding performance results. We use a compressed RFW test set in level 75 $(q=75)$ for all training strategies. Firstly, we show racially imbalanced VGGFace2 datasets training performance, which is lowest in accuracy and highest in standard deviation. A balanced BUPT-Balance dataset provides the most significant improvement in accuracy and standard deviation.
Furthermore, while compressed training imagery causes a minor decrease in standard deviation, no-chroma subsampling improves bias performance more significantly. Therefore, removing chroma sampling during compression becomes viable for reducing racial performance bias. We conclude from the abovementioned results that while compressed imagery or racially balanced training data during training improves the overall performance for all race-related categories, disparate results remain for specific phenotype characteristics. Furthermore, we highlight that the reduced retention of the chroma (colour) information affects, due to the use of chroma subsampling in lossy JPEG compression, on darker skin tones to a greater degree than on lighter skin tones. Furthermore, it is likely that the lossy image quantisation disproportionately affects finer image details on the facial region, such as those associated with monolid eye characteristics. Both areas are for further future work.


\section{Conclusion}

\noindent
This study examines the relationship between face verification performance for a given race-related phenotypic group under varying levels of lossy compressed sets. Overall, our evaluation finds that using lossy compressed facial image samples at inference time decreases performance more significantly on specific phenotypes, including dark skin tone, wide nose, curly hair, and monolid eye across all other phenotypic features. However, the use of compressed imagery during training does make the resulting models more resilient and limits the performance degradation encountered: lower performance amongst specific racially-aligned subgroups remains. Additionally, removing chroma subsampling improves FMR for specific phenotype categories more affected by lossy compression. Future work will explore the impact of lossy image quantisation across various face recognition architectures and propose corresponding results to have fair face recognition algorithms.

\noindent
\textbf{Ethical Considerations:}
This work aims to investigate the impact of lossy compression algorithms on phenotype-based racial groups from \cite{yucer2022measuring} to provide additional insight and understanding to guide the mitigation of bias in the development of future face recognition algorithms and systems. We conduct our experiments on three different face datasets publicly available for research use only. The reader is directed to the source publication and the associated research organisation for access to these datasets. 

\clearpage
{\small
\bibliographystyle{ieeesrt}
\bibliography{egbib}
}

\clearpage
\pagenumbering{gobble}

\beginsupplement
\maketitle

\balance

\section{FMRs on Selected Compression Levels}
\begin{minipage}[ht]{2\linewidth}
\begin{multicols}{2}
We provide down-selected compression levels differences (additional compression levels $(q=10,15,95)$) for each of the proposed training strategies using cross attribute pairings provided by \cite{yucer2022measuring}. 

\end{multicols}
\end{minipage}
\begin{minipage}[c]{2.05\linewidth}
\begin{figure}[H]
\begin{center}
\includegraphics[width=0.95\linewidth]{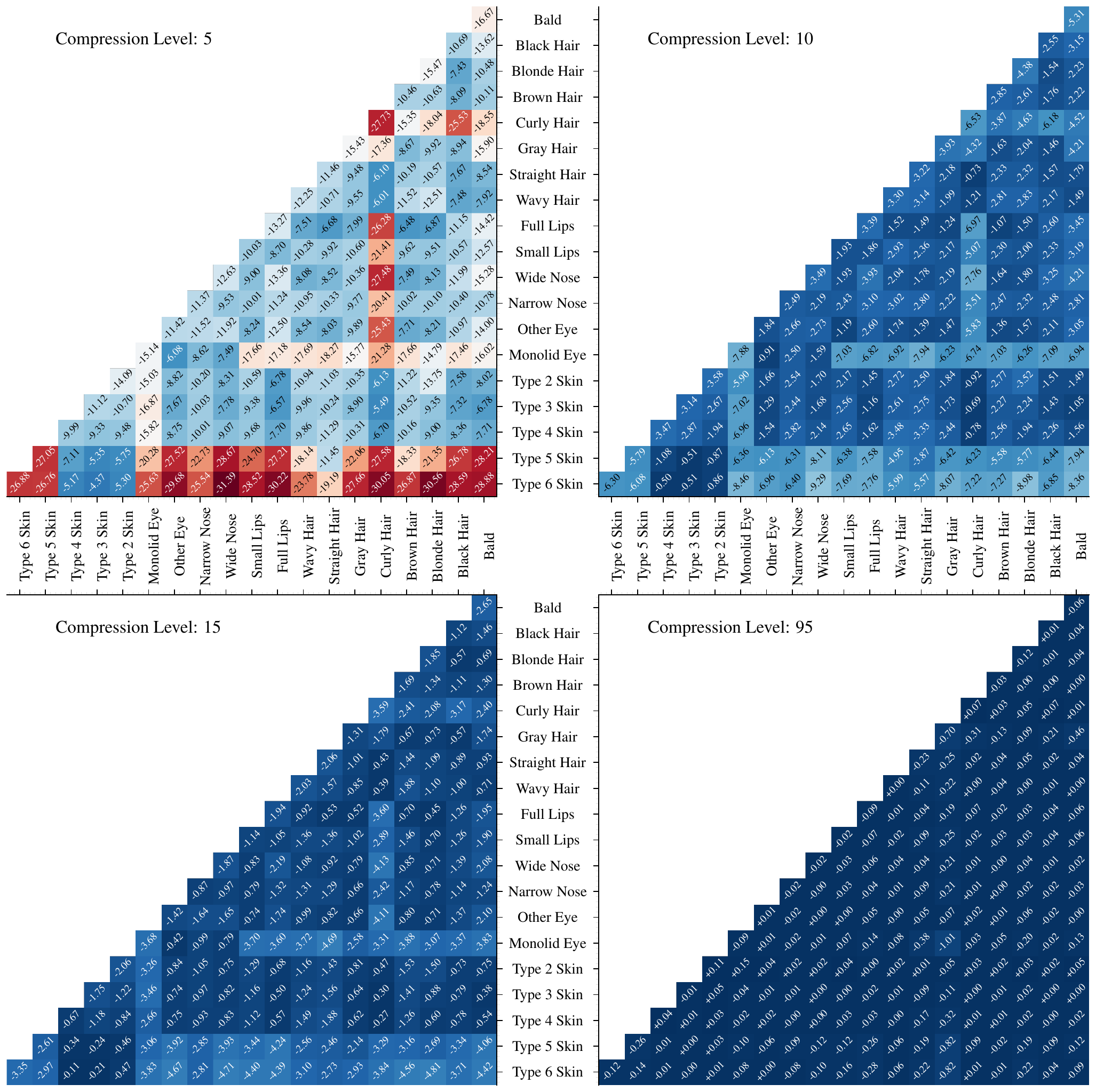}
\caption{VGGFace2 original/non-compressed training imagery and compressed RFW test imagery; FMR performance differences of cross-attribute based pairings. Each cell depicts $FMR_{original}-FMR_{q}$. \label{fig:sup_set1}}
\end{center}
\end{figure}
\end{minipage}

\newpage

\begin{figure*}[htbp]
 \centering
 \includegraphics[width=0.95\linewidth]{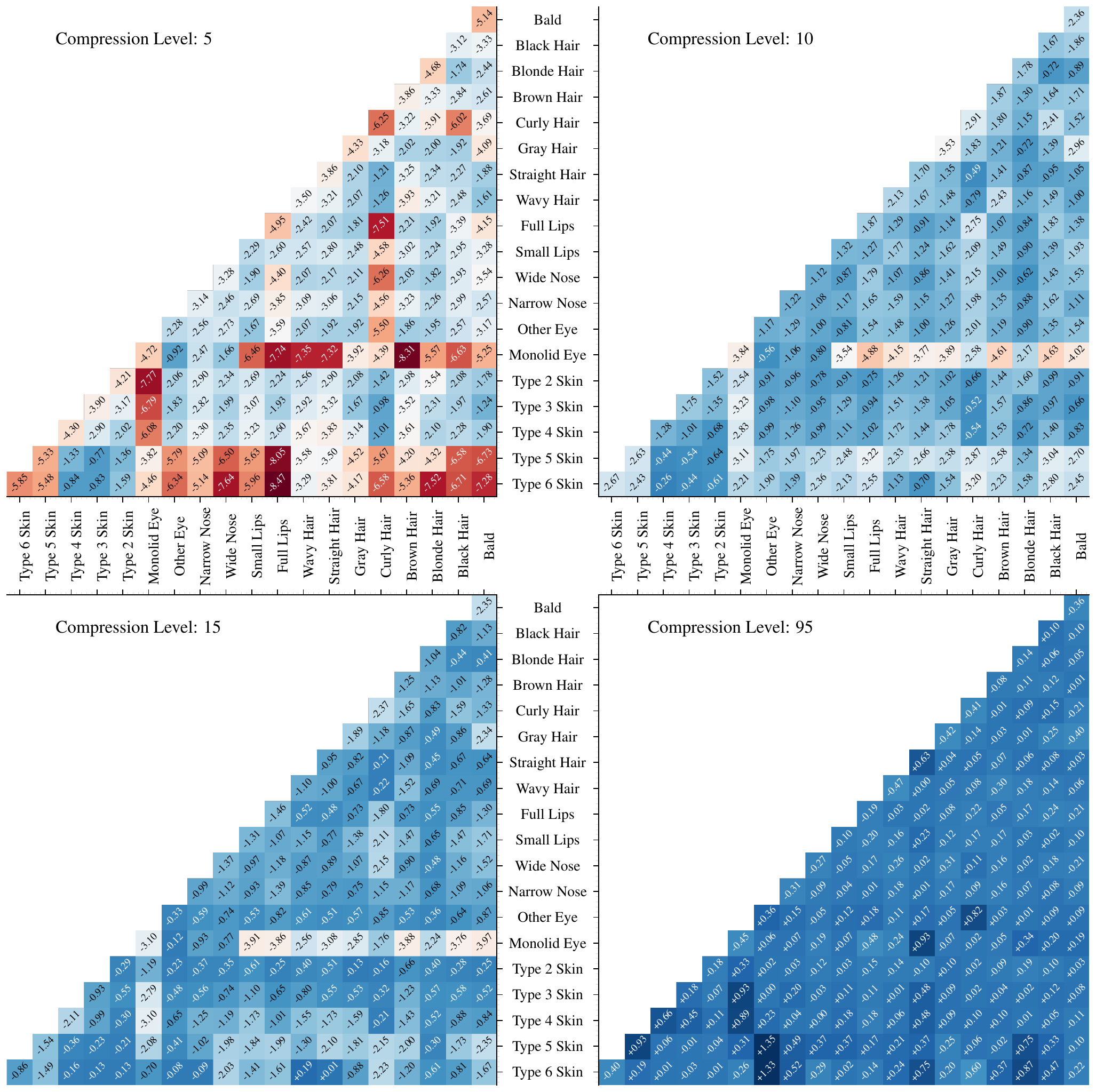}
 \caption{ VGGFace2 compressed training imagery and compressed RFW test imagery; FMR performance differences of cross-attribute based pairings. Each cell depicts $FMR_{original}-FMR_{q}$.\label{fig:sup_set3}}
\end{figure*}

\begin{figure*}[htbp]
 \centering
 \includegraphics[width=0.95\linewidth]{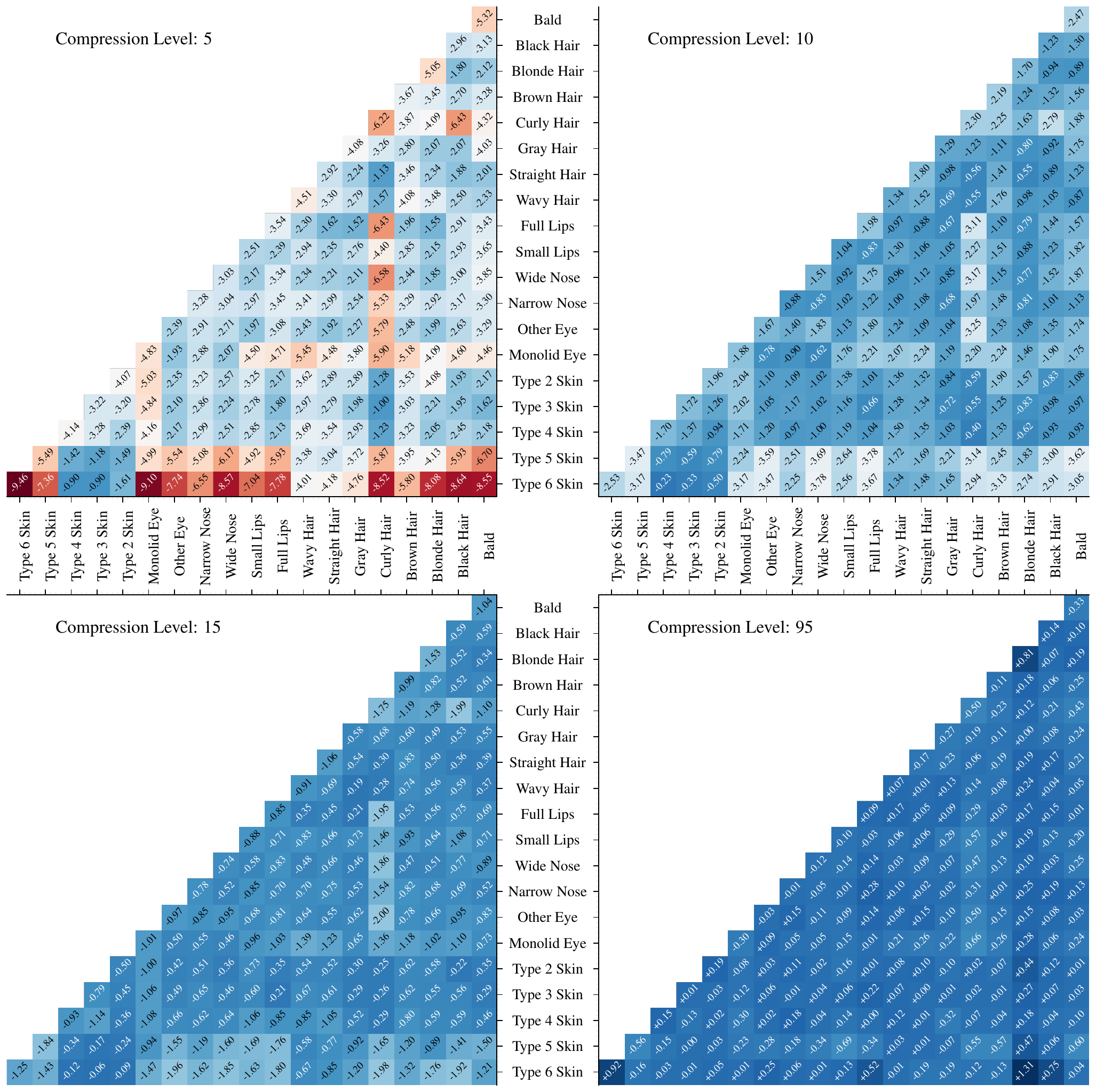}
 \caption{BUPT-Balanced compressed training imagery and compressed RFW test imagery; FMR performance differences of cross-attribute based pairings. Each cell depicts $FMR_{original}-FMR_{q}$. \label{fig:sup_set4}}
\end{figure*}

\begin{minipage}[ht]{2\linewidth}
\begin{multicols}{2}
\noindent
As described in the paper, smaller (and negative) values indicate a larger decline from the original level of performance.The FMR increases when the lossy compression increases. In Figure \ref{fig:sup_set1}, \ref{fig:sup_set2}, \ref{fig:sup_set3} and \ref{fig:sup_set4}, we demonstrate that compression level 5 (the highest compression rate) results in the most significant decrease in FMR performance for all different training strategies. In contrast, compression level 95 (the lowest compression rate) does not result in any noticeable FMR performance differences.
\end{multicols}
\end{minipage}
\begin{minipage}[c]{2.05\linewidth}
\begin{figure}[H]
\begin{center}
\includegraphics[width=0.95\linewidth]{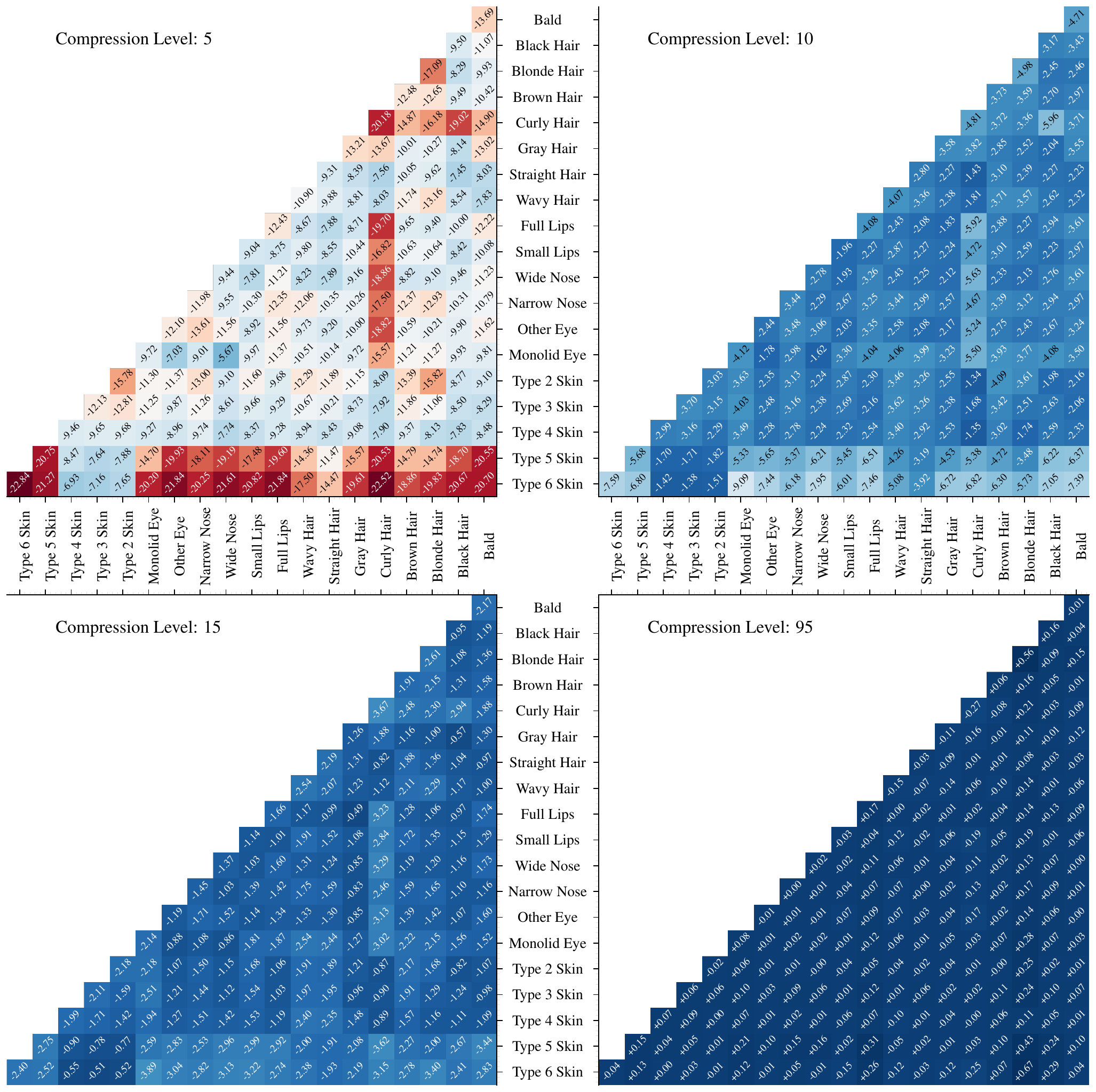}
\caption{BUPT-Balanced original/non-compressed training imagery and compressed RFW test imagery FMR performance differences of cross-attribute based pairings. Each cell depicts $FMR_{original}-FMR_{q}$.\label{fig:sup_set2}}
\end{center}
\end{figure}
\end{minipage}

\end{document}